%% file: main.tex
\documentclass[runningheads]{llncs}
\usepackage[T1]{fontenc}
\usepackage{graphicx}

\usepackage{hyperref}
\usepackage{color}

\usepackage{subcaption}
\usepackage{amsmath, amsfonts}
\usepackage{algorithm}
\usepackage[capitalise]{cleveref}
\usepackage{booktabs}
\usepackage{makecell}
\usepackage{textcomp}
\usepackage{multirow}
\usepackage{algpseudocode}

\usepackage{xcolor}
\definecolor{RoyalBlue}{RGB}{43, 101, 200}
\definecolor{BrightRed}{RGB}{187, 42, 29}

\usepackage{pifont}

\algnewcommand\algorithmicinput{\textbf{Input:}}
\algnewcommand\Input{\item[\algorithmicinput]}
\algnewcommand\algorithmicoutput{\textbf{Output:}}
\algnewcommand\Output{\item[\algorithmicoutput]}

\newcolumntype{L}[1]{>{\raggedright\let\newline\\\arraybackslash\hspace{0pt}}m{#1}}
\newcolumntype{C}[1]{>{\centering\let\newline\\\arraybackslash\hspace{0pt}}m{#1}}
\newcolumntype{R}[1]{>{\raggedleft\let\newline\\\arraybackslash\hspace{0pt}}m{#1}}

\usepackage[switch]{lineno}
\usepackage{enumitem}

\begin{document}

\title{Hear No Evil: Detecting Gradient Leakage by Malicious Servers in Federated Learning}

\author{Fei Wang\orcidID{0000-0002-1084-0690} \and
Baochun Li\orcidID{0000-0003-2404-0974}}
\institute{University of Toronto, Toronto, ON M5S 1A1, Canada \\
\email{silviafey.wang@utoronto.ca} \\ 
\email{bli@ece.toronto.edu}}

\maketitle

\begin{abstract}
  Recent work has shown that gradient updates in federated learning (FL) can unintentionally reveal sensitive information about a client's local data. This risk becomes significantly greater when a malicious server manipulates the global model to provoke information-rich updates from clients. In this paper, we adopt a defender's perspective to provide the first comprehensive analysis of malicious gradient leakage attacks and the model manipulation techniques that enable them. Our investigation reveals a core trade-off: these attacks cannot be both highly effective in reconstructing private data and sufficiently stealthy to evade detection---especially in realistic FL settings that incorporate common normalization techniques and federated averaging.

  Building on this insight, we argue that malicious gradient leakage attacks, while theoretically concerning, are inherently limited in practice and often detectable through basic monitoring. As a complementary contribution, we propose a simple, lightweight, and broadly applicable client-side detection mechanism that flags suspicious model updates before local training begins, despite the fact that such detection may not be strictly necessary in realistic FL settings. This mechanism further underscores the feasibility of defending against these attacks with minimal overhead, offering a deployable safeguard for privacy-conscious federated learning systems.

\end{abstract}


\input{intro}
\input{related}

\input{review}
\input{insights}

\input{detection}
\input{concl}

\clearpage
\bibliographystyle{splncs04}
\bibliography{main}

\appendix
\section{Appendix}

\input{manipulation.tex}

\end{document}

%% file: intro.tex

\section{Introduction}\label{sec:intro}
Federated learning (FL) has accumulated a significant amount of attention since its debut as a privacy-preserving method for distributed model training without the need to share raw data~\cite{fedavg-aistats17,fedopt-16}. However, the updates observed by the server from the clients for model aggregation can still inadvertently leak information about the raw local data, as highlighted by recent research~\cite{dlg-neurips19,idlg-20,ig-neurips20,prior-neurips21,gradinversion-cvpr21}. These gradient leakage attacks pose a serious threat to the privacy of federated learning, as they demonstrate the server's potential to reconstruct image or text data input by the global model and gradient updates, even under the assumption of honest execution of the federated learning protocol.

To enhance attack capabilities, particularly in overcoming the challenge of reconstructing data from a large batch when multiple gradient mixing steps are involved, recent research has explored a stronger threat model in which the server actively manipulates the model sent to selected clients. This manipulation induces gradient updates that reveal more information about specific data points, even in the presence of large batch sizes, multiple local update steps, or secure aggregation~\cite{robbing-iclr22,fishing-icml22,curious-eurosp23,decepticons-iclr23,gold-iclr23,hiding-iclr24,loki-sp24}. This malicious server disregards the federated learning protocol, instead altering the architecture or parameters of the model shared with the target client during specific communication rounds to stealthily obtain its private data.

The interaction between the server and the victim clients not only compromises data privacy but also undermines federated learning's training efficiency as training a maliciously modified model on its local data offers minimal contribution to the global model. Therefore, it is imperative for a selected client to promptly detect any malicious activity from the server and choose to abstain from participating in that round to prevent data leakage.
Despite the numerous studies on malicious gradient leakage attacks discussing possible ways to detect such attacks~\cite{fishing-icml22,decepticons-iclr23,curious-eurosp23}, they often assert that such detections are severely limited in the face of these attacks as they cannot accurately differentiate between model parameters affected by malicious manipulation and those resulting from legitimate optimization. Only one recent study~\cite{hiding-iclr24} has delved into client-side detectability, demonstrating that existing malicious attacks can indeed be detected by inspecting the model parameters before local training or the resulting gradients after training. 

While the study made a pioneering effort to quantify the maliciousness of models received by clients, its proposed metrics are overly simplistic and fail to capture more sophisticated threats. Their new attack, SEER, departs from prior approaches that rely on manual crafting~\cite{fishing-icml22,curious-eurosp23} or heuristic construction~\cite{robbing-iclr22}. Instead, it leverages a learning-based strategy that modifies model parameters through a joint optimization process with server-side secret decoders. As a result, neither the model parameters nor the gradients observed by the client exhibit conspicuous artifacts, making traditional detection approaches largely ineffective. Despite the theoretical appeal of SEER's approach by manipulating model weights and jointly training with secret decoders on the server side, achieving a high degree of stealth, a key practical limitation persists. Its gradient disaggregation technique critically depends on the presence of batch normalization layers, which are not only increasingly deprecated in modern neural network architectures~\cite{layernorm16,groupnorm18} but also commonly avoided in federated learning due to their adverse impact on training efficiency and stability~\cite{rethinking-distml22,flbatchrenorm-iclr24}.

In this paper, we take a defender-centric approach to examining gradient leakage attacks in federated learning under the assumption of a malicious server. Our investigation demonstrates an inherent limitation in these attacks: as they become more aggressive in extracting private data, they either inevitably leave detectable traces in the model parameters or gradients, or rely on strong assumptions about the model architecture and training setup. The remainder of this paper systematically analyzes existing attacks, identifies their practical constraints, and introduces a generalizable detection framework that can effectively safeguard client privacy. Our contributions are threefold:

First, we provide a systematic review of all notable malicious gradient leakage attacks, categorizing them based on the difficulty of detection. This comprehensive assessment reveals that the majority of model manipulation techniques either can be rendered ineffective on a technical level, even when the client does not take any explicit countermeasures, or are readily detectable through proper parameter and gradient monitoring.

Second, we uncover key insights into the limitations of more advanced, learning-based attacks that avoid manual crafting or heuristic design. Specifically, we show that their effectiveness hinges on outdated CNN architectural components and the assumption of single-step local updates on the client side---conditions intended to prevent the model from drifting away from the server's optimized manipulation for gradient decoding, but which rarely hold in practical federated learning deployments. We show that the presence of modern normalization techniques, such as layer normalization~\cite{layernorm16} or group normalization~\cite{groupnorm18}, substantially reduces the effectiveness of these attacks. Similarly, the use of federated averaging with multiple local update steps further diminishes their impact.

Third, we introduce a simple yet broadly applicable detection mechanism designed to defend against the full spectrum of known gradient leakage attacks. By performing statistical analysis on model parameters before local training begins, our detector can flag suspicious or subtle modifications, even those introduced by learning-based attacks. The framework is model-agnostic, generalizes well across architectures and datasets, adapts to diverse attack strategies, and remains lightweight and efficient, making it practical for real-world federated learning deployments.

Our findings challenge the prevailing narrative that malicious gradient leakage attacks pose an insurmountable threat to federated learning. Instead, we argue that with appropriate detection mechanisms in place, these attacks do not constitute a serious threat to the security of practical FL systems.

%% file: related.tex

\section{Related Work and Motivation}\label{sec:related}

\subsection{Gradient Leakage Attacks in Federated Learning}

Gradient leakage attacks in federated learning~\cite{fedavg-aistats17} refer to attacks where an adversary on the server side attempts to reconstruct a client's private training data using the gradients received from the client. As the name suggests, earlier works primarily focused on the FedSGD~\cite{fedsgd} setting, where each client performs a single gradient descent step and transmits gradients to the server. Under a genuine FL protocol with an honest server, these attacks~\cite{dlg-neurips19,idlg-20,gradinversion-cvpr21,ig-neurips20,prior-neurips21,tag-emnlp21,lamp-neurips22,film-neurips22,april-cvpr22,learning-uai23} can directly reveal one or more  of a client's image or text data from the gradients.

However, reconstruction faces significant limitations when the batch size is large~\cite{ig-neurips20,gradattack-neurips21}, as gradients are averaged across the mini-batch. Additionally, in more practical federated learning settings such as FedAvg~\cite{fedavg-aistats17} or its variants~\cite{fedprox-mlsys20,scaffold-icml20}, clients perform multiple steps of gradient descent in each round before transmitting the final updates---either the new local model's parameters after training or parameter deltas (\emph{i.e.,} the difference between new and previous parameters) rather than gradients---to the server. These multiple gradient descent steps, controlled by the number of epochs and batch sizes, dramatically increase gradient confusion, making it difficult to disentangle gradients associated with particular data points from the overall gradient flow, as noted by~\cite{outpost-infocom23}.

These limitations have sparked a new wave of attacks in federated learning involving malicious servers, which manipulate the global model before sending it to clients,thereby amplifying data leakage through aggregated gradients. The adversary seeks to create a malicious model $F_{W^\dagger}$, derived from the original global model $F_W$, to enhance analytic data reconstruction from the gradient updates $\nabla W^\dagger$ sent by any client. Such attacks aim to bypass protections like large batch sizes~\cite{robbing-iclr22,fishing-icml22,curious-eurosp23,decepticons-iclr23}, which typically obfuscate individual data points by averaging gradients across many samples; FedAvg~\cite{loki-sp24}, which offers inherent privacy through multiple local update steps and parameter averaging; and even secure aggregation~\cite{eluding-ccs22,hardened-eurosp23}, which cryptographically merges updates from all participating clients to prevent the server from observing individual contributions.

\begin{figure}[ht!]
    \centering
    \includegraphics[width=1\textwidth]{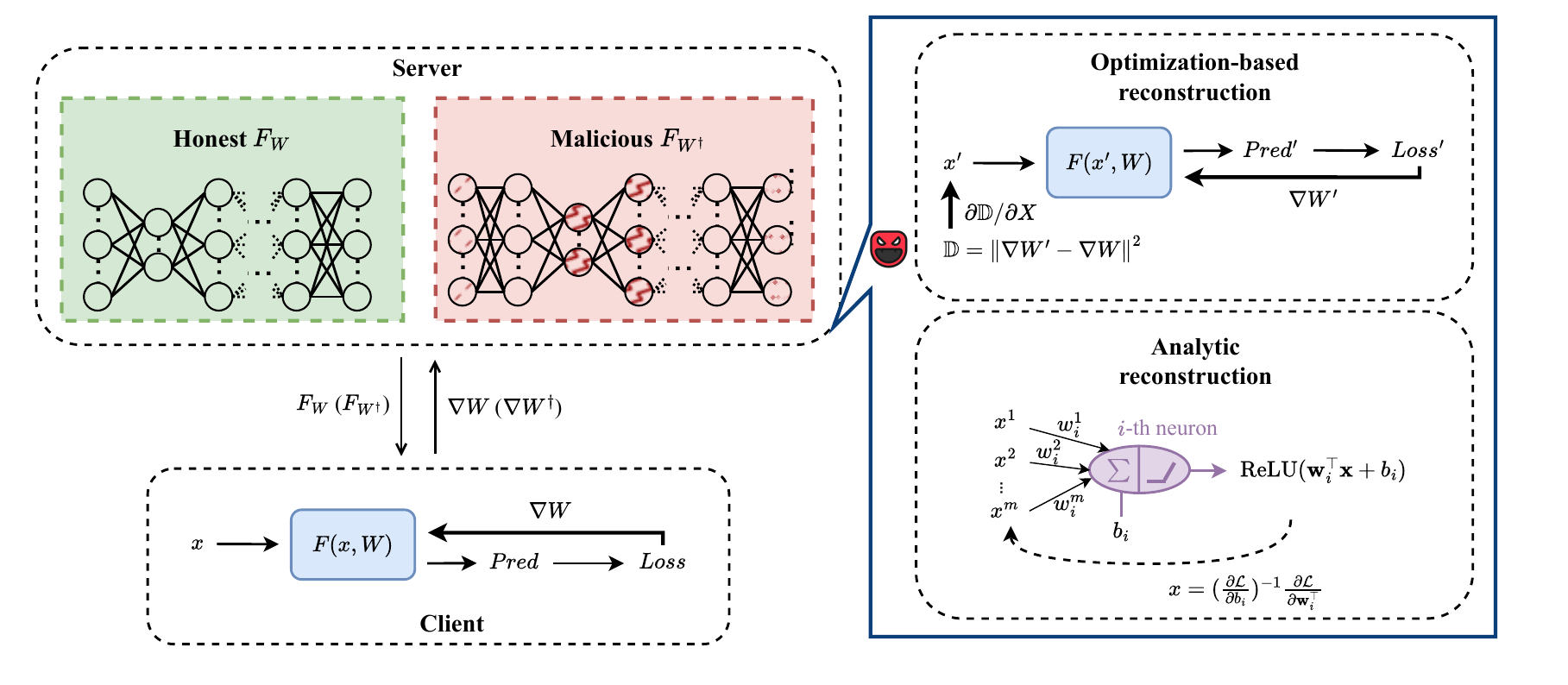}
    \caption{Honest and malicious federated learning servers executing gradient leakage attacks.}
    \label{fig:honest_malicious}
\end{figure}

Model manipulations, whether altering the architecture or modifying model parameters, aim to enhance the correlation between gradients and specific data points, thereby boosting gradient leakage at particular layers. These manipulations serve as a precursor to the final reverse-engineering step of data reconstruction from gradients, typically using one of two strategies:

\textit{Optimization-based reconstruction} iteratively optimizes dummy data to minimize the discrepancy between target gradients and those generated by the dummy input~\cite{dlg-neurips19,idlg-20,ig-neurips20,fishing-icml22}. In this approach, the gradient descent process mirrors client training but focuses on refining the dummy input while keeping the model parameters fixed.

\textit{Analytic reconstruction}, on the other hand, directly infers input data from gradients in a single step by applying closed-form solutions~\cite{robbing-iclr22,curious-eurosp23,fishing-icml22,decepticons-iclr23}. This method exploits algebraic relationships between gradients and data, often targeting the weights and biases of the first fully-connected (FC)layer (also referred to as linear layers). The success of analytic attacks hinges on the linearity of gradients with respect to the inputs. \cref{fig:honest_malicious} illustrates the gradient leakage attacks under different server assumptions in FedSGD.

\subsection{Defense or Detection of Malicious Server}
While defenses like local or distributed differential privacy~\cite{ldp-edgesys20,fldp-tifs20} and approaches tailored to defend against gradient leakage under honest servers such as~\cite{soteria-cvpr21,graddefense-infocom22,outpost-infocom23} can enhance client privacy in federated learning, strong privacy protections often come at the cost of reduced model utility~\cite{impact-neurips19}. Moreover, commonly used differential privacy has been shown to be ineffective against sophisticated malicious attacks like SEER~\cite{hiding-iclr24}, which can bypass these protections.

When a malicious server intentionally deviates from the federated learning protocol by introducing malicious model modifications that prioritize data extraction over model accuracy, it becomes unreasonable for clients to blindly trust the global model and proceed with costly local training or complex defenses. In contrast, early detection of server-side malicious activity by clients can prevent unnecessary computation, reduce the risk of exposing private information in gradient updates, and help clients avoid contributing to a compromised global aggregation.

Although malicious servers inevitably leave traces when distributing manipulated models, these adversarially modified models often embed subtle yet distinct artifacts that differentiate them from benign counterparts, presenting viable signals for detection. Nonetheless, the detection landscape remains underexplored: surprisingly little work has systematically investigated the full range of gradient leakage attacks from a detection standpoint. In this section, we offer a comprehensive review of all notable server-side gradient leakage attacks to date, organizing them by their underlying mechanisms and analyzing the corresponding detection strategies or limitations. By surfacing both the strengths and weaknesses of existing defenses, we identify a critical blind spot, particularly in detecting increasingly stealthy, handcraft-free attacks, thereby motivating our deeper investigation into this most elusive threat model.

%% file: review.tex

\section{Anatomy of Gradient Leakage by Malicious Servers in Federated Learning: Attacks and Detection}\label{sec:review}

\subsection{Rookie Moves: Model Architectural Tampering Is Easily Thwarted}\label{subsec:rookie_moves}

The success of one major stream of attacks relies on exploiting the linearity of gradients with respect to inputs, particularly when implemented through a fully connected (FC) layer followed by an activation function, especially ReLU~\cite{robbing-iclr22,curious-eurosp23,compromise-ica3pp22}. This approach requires the FC layer to be positioned as the first layer of the model, creating a direct mapping between the input data and the output space of the FC layer. However, standard CNN architectures like ResNet and VGG typically place FC layers toward the end of the network, following multiple convolutional layers for feature extraction.

To circumvent this architectural constraint, \cite{robbing-iclr22,curious-eurosp23,loki-sp24} proposed a straightforward but conspicuous solution: inserting a new FC layer at the beginning of the network, effectively bypassing the convolutional layers. While technically effective, such architectural modifications are easily detectable and lack stealth. This detection vulnerability is further exacerbated by the implementation details of popular deep learning frameworks. Specifically, when clients load model parameters from a server using functions such as \texttt{load\_state\_dict()} in PyTorch~\cite{pytorch-doc} or \texttt{load\_weights()} in TensorFlow~\cite{tensorflow-doc}, these functions enforce strict architectural consistency. If the server-provided weights contain parameters for layers not defined in the client's local model---such as the inserted FC layer---the frameworks either raise errors or silently ignore the mismatched layers, preventing the malicious modifications from taking effect. As a result, such manipulation efforts can be rendered ineffective, even when the client does not take any explicit countermeasures.

\subsection{Under the Radar: Model Parameter and Gradient Inspection}

Alternatively, to circumvent the aforementioned architectural constraint, \cite{curious-eurosp23,compromise-ica3pp22} modify the model parameters to simplify the data flow through convolutional layers, ensuring input data reaches subsequent FC layers with minimal distortion or feature extraction. Specifically, \cite{curious-eurosp23} set the convolutional kernels to (scaled) identity matrices, effectively transforming these layers into pass-through operations that preserve the spatial structure of inputs while allowing them to flow unaltered. Such modifications can be easily detected by examining convolutional filters for patterns indicating minimal feature extraction. An intuitive quick check~\cite{hiding-iclr24} is, for each filter in the first convolutional layer, to calculate the ratio of the absolute value of the largest entry to the sum of the absolute values of all other entries. Higher values indicate greater preservation of the original input.

Similarly, in Transformers for text data reconstruction, \cite{decepticons-iclr23} unmix token information across layers, keeping inputs to each layer unaltered, by disabling all attention layers and restricting the outputs of feed-forward blocks to the last entry. The query weights are set to zero, while their biases are scaled versions of the first positional encoding. The key and value weights are (partially) set to identity mappings, with all their biases set to zero. The model parameter inspection can also be extended to Transformers by examining the attention layers.

Beyond model parameter inspection, which detects malicious tampering prior to a client's local training, \cite{hiding-iclr24} pioneered the study of client-side detectability by introducing the disaggregation signal-to-noise ratio (D-SNR) for gradient inspection. This metric is formally defined as:
\begin{equation}
\text{D-SNR}(\boldsymbol{\theta}) = \max_{W \in \boldsymbol{\theta}_{lw}} \frac{\max_{i\in\{1,...,B\}} \left\|\frac{\partial \ell(F(x_i),y_i)}{\partial W}\right\|}{\sum_{i=1}^{B} \left\|\frac{\partial \ell(F(x_i),y_i)}{\partial W}\right\| - \max_{i\in\{1,...,B\}} \left\|\frac{\partial \ell(F(x_i),y_i)}{\partial W}\right\|},
\label{eq:d-snr}
\end{equation}
where $B$ denotes the batch size and $\boldsymbol{\theta}_{lw}$ represents the set of weights of all FC and convolutional layers. The D-SNR metric effectively identifies dominant gradients within batch gradients for each layer. When one sample's gradient norm substantially exceeds others in the batch, D-SNR values increase significantly. This detection capability is particularly effective against attacks such as~\cite{fishing-icml22,eluding-ccs22,hardened-eurosp23}, which manipulate the weights and biases of the final classification layer to isolate gradient updates associated with specific classes or features.

\begin{figure}[ht!]
	\centering
	\begin{subfigure}[b]{0.35\textwidth}
		\centering
		\includegraphics[width=\textwidth]{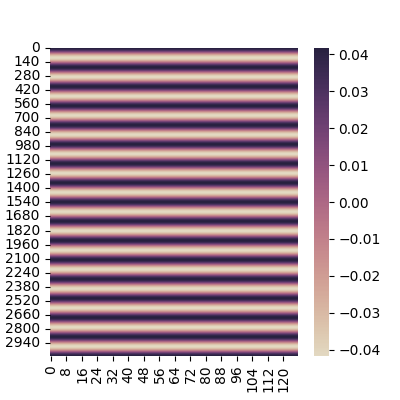}
		\caption{Weight matrix}
		\label{fig:fc_weight}
	\end{subfigure}
	\begin{subfigure}[b]{0.35\textwidth}
		\centering
		\includegraphics[width=\textwidth]{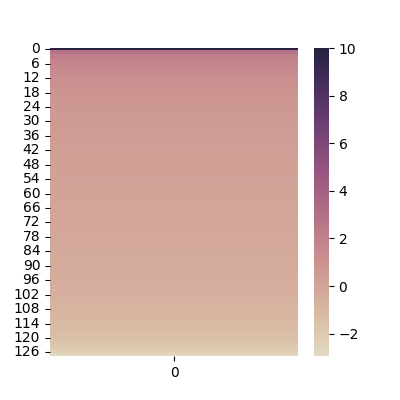}
		\caption{Bias matrix}
		\label{fig:fc_bias}
	\end{subfigure}

    \begin{subfigure}[b]{0.33\textwidth}
		\centering
		\includegraphics[width=\textwidth]{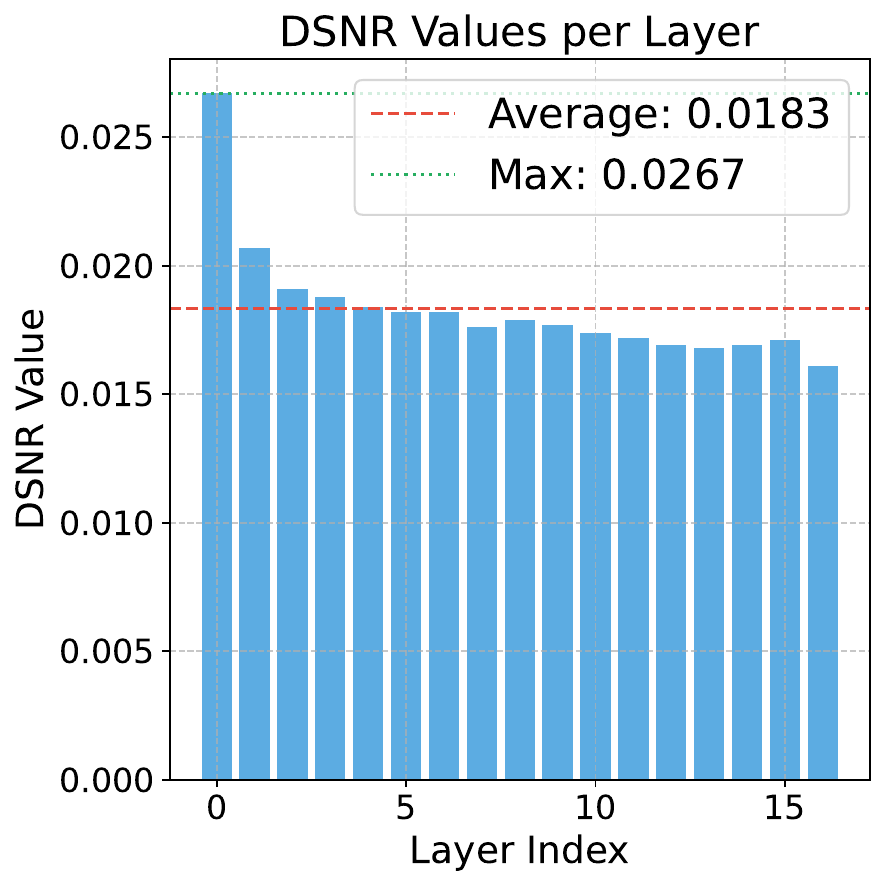}
		\caption{D-SNR before attack}
		\label{fig:dsnr_before_robbing}
	\end{subfigure}
    \begin{subfigure}[b]{0.33\textwidth}
		\centering
		\includegraphics[width=\textwidth]{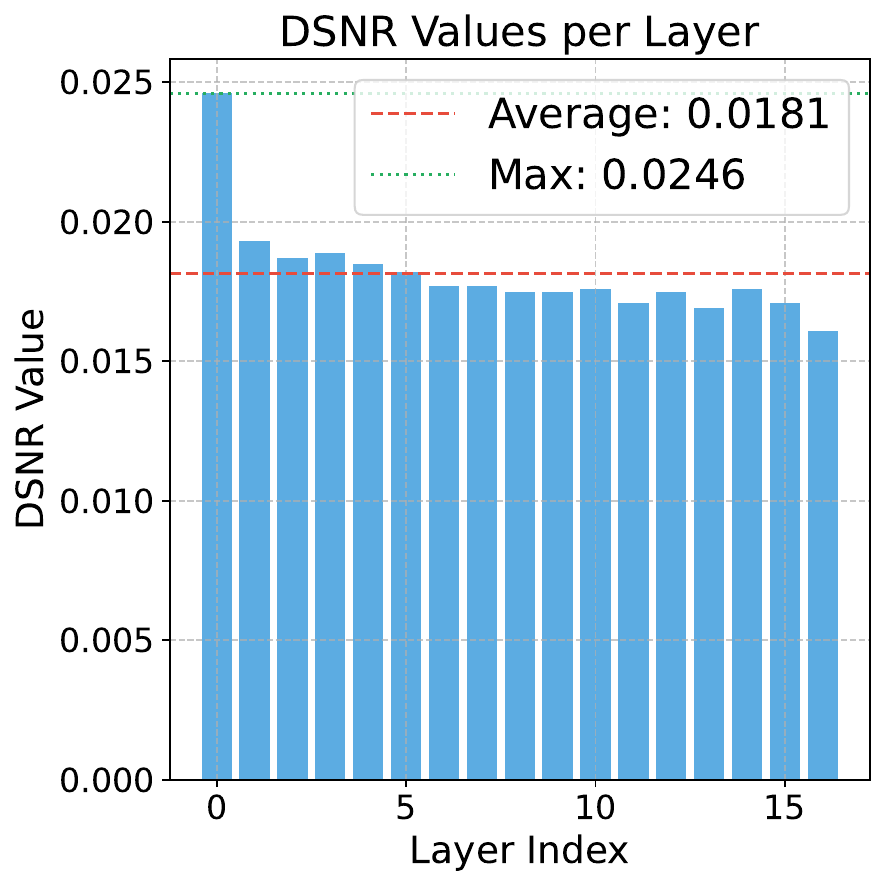}
		\caption{D-SNR after attack}
		\label{fig:dsnr_after_robbing}
	\end{subfigure}%

	\caption{The weights and bias of the imprint layer in ResNet-18 under the attack from~\cite{robbing-iclr22}, along with the D-SNR values for each monitored layer before and after the attack.
    }
	\label{fig:robbing_fc}
\end{figure}

However, we find that the D-SNR fails to detect the maliciousness of the attacks in~\cite{curious-eurosp23,robbing-iclr22}, despite the fact that the modifications to the imprint linear layer are highly conspicuous. As illustrated in the following toy example, we conduct the attack from~\cite{robbing-iclr22} on ResNet-18 using the default hyperparameters provided in their implementation. The number of bins is set to 128, and the linear function used to configure the weight is the Discrete Cosine Transform (DCT-II) coefficient~\cite{dct-itip20}. 
In~\cref{fig:robbing_fc}, we visualize the values of the weight and bias of the imprint linear layer. The dimension of the imprint layer is determined by the input feature dimension and the number of bins. In this particular case, the size of the weight matrix is $(3 * 32 * 32) \times 128$, \emph{i.e.,} $3072 \times 128$. As observed from~\cref{fig:fc_weight}, for each row (input feature), the weights across all columns (bins) have the same value. Additionally, a repeating pattern occurs every $192$ (\emph{i.e.,} $3 * 32 * 2$) consecutive rows, where the weights gradually decrease from the largest value to the smallest and then symmetrically increase back to the largest value. In~\cref{fig:fc_bias}, it is shown that the bias of each bin monotonically decreases, except for the very first one.

Nevertheless, the D-SNR metric for each layer remains similar before and after the attack, with values close to 0 rather than exceeding the malicious threshold of 1 (or infinity for the attack in~\cite{fishing-icml22}), as shown in~\cref{fig:dsnr_before_robbing,fig:dsnr_after_robbing}. This can be attributed to the fact that the modifications to the weights and bias of the imprint linear layer are intended to facilitate analytic reconstruction of multiple data points on the first FC layer, rather than disaggregate the average gradient for a single data point. As a result, the D-SNR fails to detect any dominant gradients. Alternatively, it may not account for the imprint layer inserted as the first FC layer after the architectural modification.

\subsection{The Invisible Hand: When Inspection Fails Against Handcraft-Free Manipulations}

While examining model parameters and gradients can reveal obvious, handcrafted tampering, more subtle and inconspicuous manipulations remain difficult to detect. A notable example is SEER~\cite{hiding-iclr24}, which modifies model parameters through SGD optimization prior to FL deployment while simultaneously training a secret disaggregator and reconstructor using auxiliary data assumed to be available at the server. This optimization process specifically alters the objective function to favor data reconstruction capabilities over model performance on the original task, ensuring the variance of gradient norms remains small, which directly results in low D-SNR values comparable to legitimate models. Despite this evasion, SEER can still reconstruct an image satisfying a selected property from an arbitrary mini-batch of client data, even with very large batch sizes such as 512, by employing its trained secret disaggregator and reconstructor components~\cite{hiding-iclr24}.

Such learning-based attacks are particularly difficult to detect through statistical analysis of model parameters or gradients, as the malicious modifications are introduced naturally during the optimization process rather than through manual intervention, making them hard to distinguish from normal training updates. \cite{hiding-iclr24} claims that SEER is currently the only gradient leakage attack by malicious servers that is effective, stealthy, and undetectable at the same time. However, before moving to client-side detectability against SEER and other attacks that leverage the natural optimization process to modify model parameters, we first provide key insights into their practicality, the underlying ``secret sauce'' they depend on, and how these assumptions can be readily nullified by realistic federated learning systems.

%% file: insights.tex

\section{Breaking Handcraft-Free Attacks with Modern Models and Federated Learning}\label{sec:insights}

Among various learning-based attacks, we focus on the SEER attack~\cite{hiding-iclr24} in this section, as it does not require architectural modifications to the model and cannot be easily rejected by the client using straightforward verification mechanisms discussed in~\cref{subsec:rookie_moves}.

\subsection{Enhanced Feature Fishing via Batch Normalization}

The success of the feature fishing attack~\cite{fishing-icml22} can be attributed to its exploitation of the feature distribution properties within the batch. If the attacker knows or accurately estimates the cumulative distribution function (CDF), assumed to be continuous, for some feature of interest corresponding to the target class, then the attacker can partition the feature space into ``bins'' of equal probability mass, each with mass $\frac{1}{M}$ (where $M$ should be larger than the batch size $B$~\cite{robbing-iclr22}). Let $X$ denote the random variable representing the number of data points from the target class in the batch that fall into the selected bin. 

The probability of having exactly one data point from the target class in the bin can be derived as follows:
\begin{align}
    p(X = 1)
    &= \sum_{i=1}^{M} P(x_i \in \text{bin} \land \forall j \neq i, x_j \notin \text{bin})
    \\
    &= \binom{M}{1} \cdot P(x_1 \in \text{bin} \land \forall j \neq 1, x_j \notin \text{bin})
    \\
    &= \binom{M}{1} \cdot P(x_1 \in \text{bin})
    \cdot P(\forall j \neq 1, x_j \notin \text{bin} \mid x_1 \in \text{bin})\\
    &= \binom{M}{1} \cdot \frac{1}{M} \cdot \left(1 - \frac{1}{M}\right)^{M-1} = \left(1 - \frac{1}{M}\right)^{M-1}
\end{align}

This derivation relies on the assumption that samples' bin membership is independent and uniformly distributed, with each target sample falling into the bin with probability $\frac{1}{M}$. As $M$ increases, this probability approaches a lower bound: $\frac{1}{e} \approx 0.3679$.

Building on these insights, SEER~\cite{hiding-iclr24} claims to enhance the lower bound by exploiting batch normalization (BN) layers~\cite{batchnorm15} in DenseNet~\cite{densenet} and ResNet~\cite{resnet} models. However, while their paper downplays this aspect, we will demonstrate that the presence of BN is, in fact, a crucial factor for successfully stealing a single image, even from batches as large as 512. By normalizing inputs across the batch in a channel-wise manner, BN introduces \textbf{inter-sample dependencies}, directly violating the independence assumption that forms the foundation of the original lower bound. 

In networks with batch normalization, the computational graphs of different samples in a batch become coupled, as each sample's forward pass (and consequently its gradient) depends on the statistics computed over the entire batch at each batch normalization layer. Formally, a batch normalization layer normalizes activations according to:
\begin{equation}
\hat{h}_i = \frac{h_i - \mu_B}{\sqrt{\sigma_B^2 + \epsilon}}
\end{equation}
where $h_i$ represents pre-normalization activations for sample $i$, $\mu_B = \frac{1}{B}\sum_{j=1}^{B}h_j$ is the batch mean, and $\sigma_B^2 = \frac{1}{B}\sum_{j=1}^{B}(h_j - \mu_B)^2$ is the batch variance. 

The gradient of each sample with respect to model parameters depends on all other samples:
\begin{equation}
\nabla_\theta \mathcal{L}(x_i) = \underbrace{\frac{\partial \mathcal{L}}{\partial \hat{h}_i}\frac{\partial \hat{h}_i}{\partial h_i}\frac{\partial h_i}{\partial \theta}}_{\text{direct effect}} + \underbrace{\sum_{j \neq i}\frac{\partial \mathcal{L}}{\partial \hat{h}_j}\frac{\partial \hat{h}_j}{\partial h_i}\frac{\partial h_i}{\partial \theta}}_{\text{indirect effect via batch statistics}}
\label{eq:gradient_bn}
\end{equation}

These dependencies allow SEER to define properties based on the local in-batch distribution rather than global distribution. Specifically, instead of using pre-defined global bins, they can target samples satisfying specific relative properties within each batch, such as the sample with the minimum (or maximum) value of a chosen feature $m$, e.g., image brightness, red channel value, and etc., $\mathcal{P}_{\text{min}}(x) = \mathbf{1}[m(x) = \min_{j \in [B]} m(x_j)]$. For example, if we choose $m$ to be the brightness of the image, then $\mathcal{P}_{\text{min}}(x)$ is 1 if $x$ is the darkest image in the batch.

With this local property, the probability calculation changes fundamentally. The event of selecting a specific target sample now depends on its relative position within the batch distribution. Let's derive the new probability lowerbound. When selecting the sample with the minimum value of feature $m$ within the batch:
\begin{align}
    P(x_i \in \text{prop})  &= P(m(x_i) = \min_{j \in [B]} m(x_j))\\
&= P(m(x_i) < m(x_j) \text{ for all } j \neq i)
\end{align}

Let $p_j = P(m(x_j) < m(x_1))$ denote the probability that sample $x_j$ has a lower feature value than the target sample $x_1$ (without loss of generality, we assume $x_1$ is the target), then the probability that exactly one sample satisfies the selection property becomes:
\begin{align}
p(X' = 1) &= P(x_i \in \text{prop}) \cdot P(\forall j \neq i, x_j \notin \text{prop} \mid x_i \in \text{prop}) \\
&= \prod_{j \neq i} P(m(x_i) < m(x_j)) \cdot 1 = \prod_{j \neq i} (1 - p_j).
\end{align}
This follows directly from the construction of the minimum property $\mathcal{P}_{\text{min}}(x)$, which selects exactly one sample per batch. Thus, once the target sample is selected as the minimum, no other sample can be, leading to $P(\forall j \neq i, x_j \notin \text{prop} \mid x_i \in \text{prop}) = 1$.

Batch normalization, under an attacker's manipulation, can facilitate sharper feature separation across samples. When combined with careful feature selection, this can drive the probabilities $p_j$ toward zero in the optimal case, resulting in:
\begin{equation}
p(X' = 1) \approx 1 - \epsilon' 
\end{equation}
where $\epsilon'$ is small error term capturing potential ties or numerical imprecisions. This explains why SEER empirically achieves $p(X = 1) > 0.9$ even with batch sizes as large as 512~\cite{hiding-iclr24}, significantly outperforming the $\frac{1}{e} \approx 0.3679$ bound of approaches that don't leverage batch normalization.

\subsubsection{Batch normalization is outdated in modern models}
While SEER does not require direct access to batch normalization statistics, it still relies on the presence of batch normalization layers within the model, enabling the attacker to manipulate the model weights to achieve the desired feature separation. Although the distinct gradient signal from the target sample only becomes apparent to the attacker after decoding via the secret disaggregator, remaining highly stealthy from the client's perspective, this reliance on batch normalization severely limits SEER's practicality in modern architectures that have largely abandoned batch normalization in favor of alternative normalization techniques. 

Since around 2016, the field has seen a steady shift away from batch normalization toward alternative normalization techniques. Modern convolutional neural networks predominantly employ alternative methods such as layer normalization (LN)~\cite{layernorm16}, which is dominant in Vision Transformers~\cite{transformers-emnlp20,vit-iclr21} or other transformer-based models~\cite{bert,gpt2,gpt4}, group normalization (GN)~\cite{groupnorm18}, commonly used in modern vision backbones such as certain ResNet and Faster R-CNN variants, and instance normalization~\cite{instancenorm16}, which, while more niche, remains notable in specific applications. Crucially, these techniques differ fundamentally from batch normalization in one important aspect: \textbf{they normalize in a sample-wise manner and do not introduce dependencies between individual samples within a batch}. For example, layer normalization normalizes activations across features for each individual sample independently, while group normalization operates within groups of channels, also independently for each sample. In the absence of inter-sample dependencies introduced by batch normalization, the gradient of each sample in~\cref{eq:gradient_bn} can be simplified to just the first term.


The local in-batch distribution properties exploited by SEER are no longer relevant in modern architectures. This shift effectively nullifies SEER's core mechanism, rendering it ineffective against the majority of contemporary networks. Consequently, for these architectures, the fundamental assumption of sample independence remains valid, and the theoretical lower bound cannot be improved beyond the original $\frac{1}{e}$ limit.

\subsubsection{Batch normalization is not favorable for federated learning}
It is known that BN can cause accuracy degradation in certain FL settings, particularly when using small or non-IID mini-batches~\cite{batchrenorm-nips17,groupnorm18}. The core challenge with BN in FL stems from the fact that channel-wise statistics are trained locally and become inconsistent across devices. When these models are aggregated, the mismatched statistics can degrade the performance of the global model. \cite{rethinking-distml22} demonstrates that layer normalization offers superior performance in FL by effectively mitigating external covariate shift. Unlike BN, LN computes normalization statistics independently for each sample, eliminating inter-client statistical variance. As a result, normalization becomes independent of batch composition and size, ensuring consistent behavior across heterogeneous client data distributions.

\subsubsection{Federated averaging inherently neutralizes learning-based model manipulations}

While newer approaches like~\cite{loki-sp24} claim to work with FedAvg by enabling data reconstruction from model updates after multiple local iterations (rather than simple gradients as in FedSGD), they rely on architecture-specific modifications and parameter alterations through convolutional layers. As previously established, such attacks are readily detectable through parameter inspection. On the other hand, learning-based model manipulations like SEER, despite being inconspicuous in parameter and gradient space, face a fundamental obstacle in the FedAvg setting.

The FedAvg algorithm~\cite{fedavg-aistats17} inherently counteracts such manipulations through its core mechanism: clients perform multiple local epochs of training across numerous batches before transmitting updates to the server~\cite{outpost-infocom23}. This process naturally overwrites the malicious server's crafted parameter manipulations, as clients optimize for a fundamentally different objective, e.g.,minimizing classification loss, whereas the server aims to enable attacks such as gradient disaggregation or data reconstruction. Any adversarial optimization necessarily diverges from the client's task-driven objective.

Consider SEER's optimization objective, $\mathcal{L}_{\text{SEER}} = \left\| r\left(d(\mathbf{g}_{\text{rec}})\right) - \mathbf{x}_{\text{rec}} \right\|_2^2
+ \alpha \cdot \sum_{i \in I_{\text{nul}}} \left\| d(\mathbf{g}_i) \right\|_2^2
$, where $d$ is the disaggregator and $r$ is the reconstructor, which contrasts with the client's sole focus on minimizing $\frac{1}{N} \sum_{i=1}^{N} -\hat{y}_i \log(m(x_i))$, where $m$ represents the shared model in FL. This misalignment leads to elevated initial losses during client training with a manipulated model. As clients perform multiple SGD steps, they naturally overwrite the server's manipulations, steering the model parameters toward minimizing only the task loss.

This fundamental tension between server manipulation and client optimization creates an insurmountable barrier to the effectiveness of attacks like SEER in practical federated learning settings. Even with as few as 5 communication rounds of FedSGD, the attack's effectiveness degrades significantly~\cite{hiding-iclr24}---a problem that becomes exponentially more severe when multiple local epochs over several batches are performed in FedAvg. The attack's reliance on preserving specific parameter configurations becomes untenable when clients perform substantial local training, which is standard practice in FedAvg deployments. Consequently, any learning-based model manipulation strategy from the server becomes largely ineffective once clients apply their own optimization steps guided by the original learning objective.

%% file: detection.tex

\section{A Simple yet General-Purpose Anomaly Detection Mechanism}\label{sec:detection}
Despite the readily detectable nature of attacks that modify model parameters with particular patterns (either through manual changes or construction functions), and the impracticality of handcraft-free learning-based attacks that modify model parameters through natural optimization, we propose a simple yet general-purpose anomaly detection mechanism. This mechanism can be efficiently performed on client devices to protect FL clients in any vulnerable situation, even when assuming the client accepts any architecture changes, serving as a comprehensive defense against malicious gradient leakage attacks from adversarial servers.

\subsection{The Workflow and Design Principles}

To efficiently detect a wide range of malicious gradient leakage attacks, we introduce a \textbf{``warm-up phase''} before the onset of regular local training in each communication round, immediately after receiving model parameters from the server. This phase enables the client to collect essential statistics for detection by performing a few iterations of gradient descent on a small, randomly selected subset of its local training data. In each iteration, the client captures snapshots of the updated model and monitors relevant changes for anomaly analysis. At the end of the warm-up phase, based on the detection outcome, benign or malicious, the client can decide either to proceed with local training or abstain from it, withholding model updates for that communicationround.

Our detector is designed to be simple yet general-purpose, meeting the following key requirements: (1) \textit{model-agnostic}: it can operate across arbitrary neural network architectures and structural nuances; (2) \textit{attack-adaptive}: it generalizes across any potential hyperparameter settings of an arbitrary attack; (3) \textit{lightweight and efficient}: it can be efficiently deployed on resource-constrained client devices, without incurring significant computational or memory overhead, or delay to the main learning task.

\subsection{Statistical Profiling of Targeted Layers}\label{subsec:layer}

To ensure compatibility with a wide range of model architectures regardless of depth or width, we focus on monitoring only the layers most vulnerable to malicious manipulation, rather than analyzing all layers. As discussed in~\cref{sec:review}, optimization-based and analytic data reconstruction attacks consistently target a few key components: \textit{the first and final FC layers}, \textit{convolutional layers} (especially in CNNs), and \textit{attention layers} (particularly in Transformers), with further details in~\cref{subsec:manipulation}. Additionally, learning-based attacks such as SEER exploit \textit{batch normalization layers}. These layers are thus the inevitable targets across various attack strategies. To improve efficiency, we focus on the first and last FC layers, the first convolutional layer if present, up to three batch normalization layers, and the first and last attention layers if the model is a Transformer.

For most of the target layers, we monitor the distributions of weight and bias parameters. For batch normalization layers, we track their activations, as a targeted detection against attacks like SEER. Although such attacks introduce only subtle changes to model parameters through natural optimization, they can still produce abnormal activation patterns, especially noticeable across iterations during our warm-up phase. For each target layer, our detector monitors and tracks four key statistical measures derived from the weight and bias tensors, as well as the activations after batch normalization:
\begin{enumerate}[label=\textbullet]
    \item \textbf{Mean value:} to detect systematic shifts in the central tendency of parameter distributions;
    \item \textbf{Standard deviation:} to capture changes in the variability or spread of parameters;
    \item \textbf{Maximum absolute value:} to identify extreme outliers or abnormal scaling patterns.
\end{enumerate}

We maintain a historical record of these statistics throughout the warm-up phase, creating a statistical profile for the model after each iteration's update. During detection, the current statistics are compared against this historical baseline using Z-score analysis~\cite{zscore-ijca18}. For each monitored metric, the Z-score $Z = \frac{x - \mu}{\sigma}$ is computed to quantify the deviation of the current value from its historical mean. Statistically significant deviations, manifested as large Z-scores, indicate anomalous behavior potentially caused by adversarial manipulations. By monitoring the Z-score distributions across target layers or activations, this approach enables detection of subtle statistical signatures of various attacks without requiring prior knowledge of specific attack implementations or hyperparameter settings. If the Z-score of any monitored metric in any target layer exceeds a predefined threshold, the model is flagged as malicious, indicating it has likely been manipulated for data reconstruction or is vulnerable to gradient leakage. Otherwise, the model is considered benign.

This multi-step warm-up phase eliminates the need to inspect gradients after just one iteration of gradient descent, as the gradient updates are reflected in the model parameters. It also helps track the magnitude and velocity of changes in target metrics as the client's optimization diverges from the model trained with SEER.

\subsection{Performance Evaluation}

Our implementation and evaluation are based on the open-source codebases Breaching~\footnote{https://github.com/JonasGeiping/breaching} and Plato~\footnote{https://github.com/TL-System/plato}. Breaching provides a comprehensive suite of gradient leakage attacks by malicious servers, while Plato offers an extensible framework for federated learning. Experiments were conducted on a machine with an Intel i7-13700K CPU, 128GB DDR5 RAM, an NVIDIA RTX 4090 GPU (24GB), running Ubuntu 22.04 and CUDA 12.3. For the detection settings, we perform 5 warm-up steps, \emph{i.e.,} 5 iterations of gradient descent on a randomly selected subset of 128 samples from the local training data. The detection process is efficient, taking only 0.72 to 0.88 seconds from the start to the end of the warm-up phase. This duration is negligible compared to the time required for local training and is also comparable to the time needed for computing the D-SNR metric.

\begin{figure}[ht!]
	\centering
	\begin{subfigure}[b]{0.4\textwidth}
		\centering
		\includegraphics[width=\textwidth]{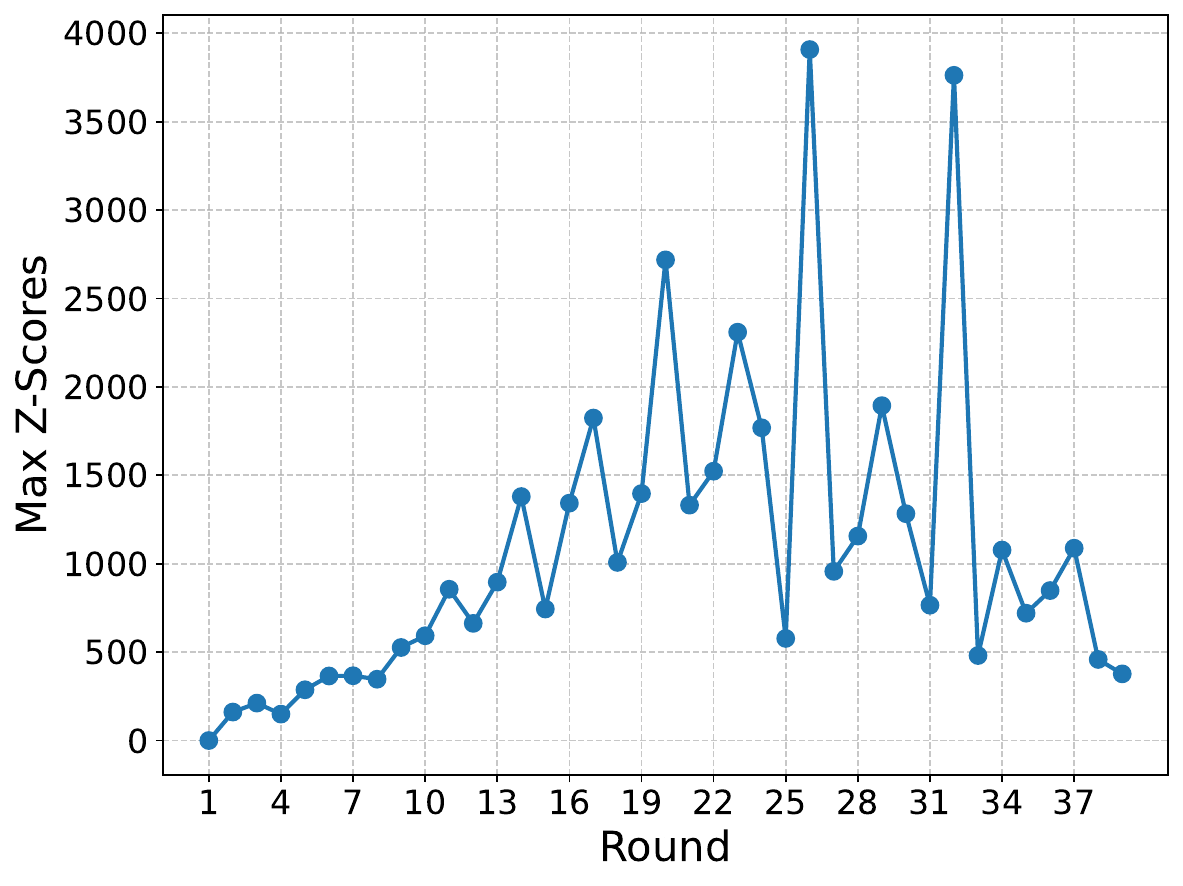}
		\caption{Maximum Z-scores}
		\label{fig:benign}
	\end{subfigure}%
	\begin{subfigure}[b]{0.4\textwidth}
		\centering
		\includegraphics[width=\textwidth]{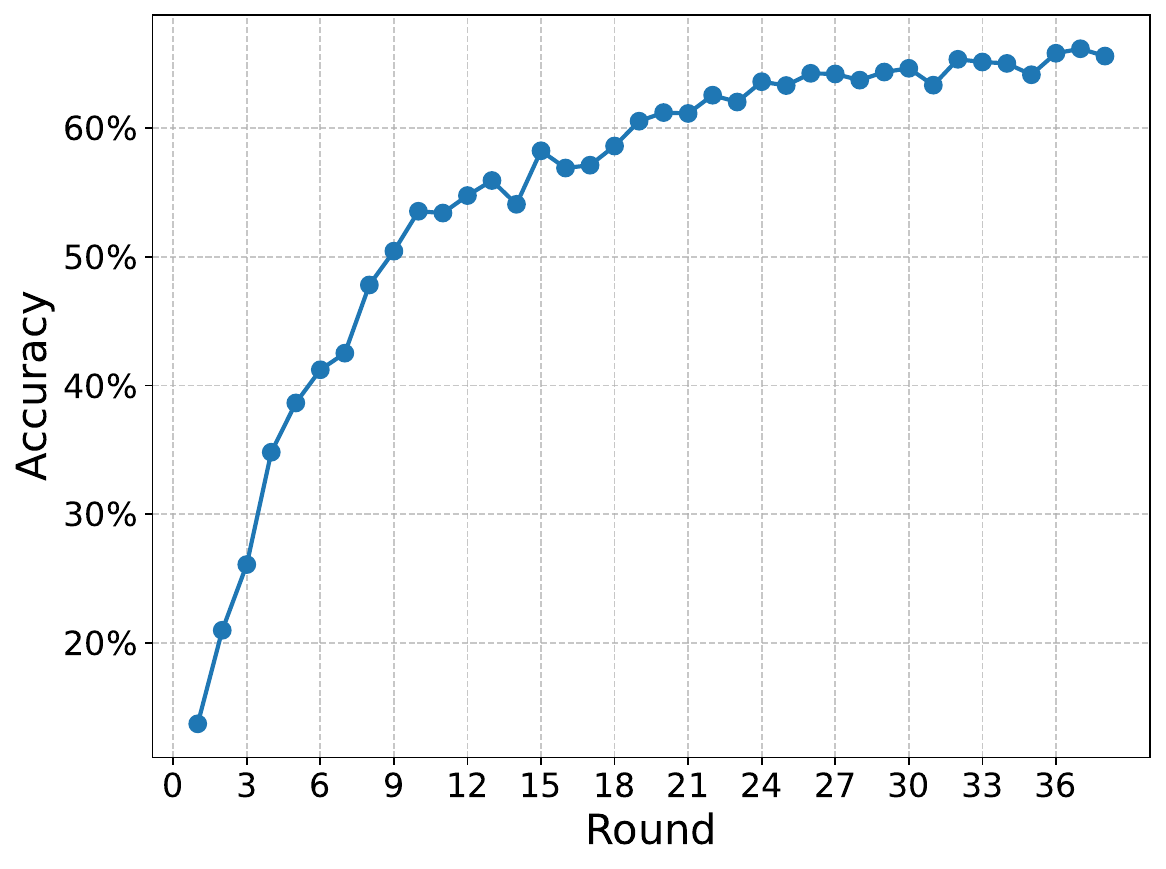}
		\caption{Accuracy}
		\label{fig:acc}
	\end{subfigure}
	\caption{A benign ResNet18 model in a natural training process.}
	\label{fig:natural}
\end{figure}

\begin{table}[ht!]
    \centering
    \caption{Detection results against the attack proposed in~\cite{fishing-icml22}. For comparison, the maximum D-SNR score increases from 0.0271 before the attack to infinity after the attack.}
    \begin{tabular}{lccc}
    \toprule
    \textbf{Target Layer} & \textbf{Z-mean} & \textbf{Z-std} & \textbf{Z-max} \\
    \midrule
    \texttt{conv[0]\_weight}         & \(1.45 \times 10^0\) & \(3.25 \times 10^1\) & \(4.82 \times 10^1\) \\
    \texttt{linear[0]\_weight}       & \(\mathbf{4.87 \times 10^5}\) & \(\mathbf{2.16 \times 10^3}\) & \(\mathbf{8.14 \times 10^2}\) \\
    \texttt{linear[0]\_bias}         & \(\mathbf{9.75 \times 10^{10}}\) & \(\mathbf{1.07 \times 10^7}\) & \(\mathbf{1.57 \times 10^6}\) \\
    \texttt{linear[-1]\_weight}      & \(\mathbf{4.87 \times 10^5}\) & \(\mathbf{2.16 \times 10^3}\) & \(\mathbf{8.14 \times 10^2}\) \\
    \texttt{linear[-1]\_bias}        & \(\mathbf{9.75 \times 10^{10}}\) & \(\mathbf{1.07 \times 10^7}\) & \(\mathbf{1.57 \times 10^6}\) \\
    \texttt{batchnorm[0]\_activation}     & \(2.80 \times 10^0\)  & \(6.30 \times 10^{-1}\) & \(8.70 \times 10^{-1}\) \\
    \texttt{batchnorm[1]\_activation}     & \(4.60 \times 10^{-1}\) & \(4.69 \times 10^0\)  & \(2.85 \times 10^0\)  \\
    \texttt{batchnorm[2]\_activation}     & \(2.04 \times 10^1\)  & \(4.62 \times 10^0\)  & \(7.00 \times 10^{-2}\) \\
    \bottomrule
    \end{tabular}
    \label{tab:fishing}
\end{table}

\cref{fig:natural} shows the maximum Z-scores of a benign ResNet18 model monitored in the warm-up phase during multiple rounds in a natural centralized training process and the accuracy of the model over rounds. As training progresses and model accuracy improves prior to the 18th round, the maximum Z-scores exhibit a gradual upward trend with minor fluctuations. After the 18th round, when the model's convergence rate slows, the Z-scores begin to fluctuate more noticeably, but still remain within a bounded range, with peak values staying below 4000.

In contrast, we show in~\cref{tab:fishing} the statistical profiles of the target layers of ResNet18 trained on CIFAR100 under the attack proposed in~\cite{fishing-icml22} after just one warm-up step. Notably, the Z-mean, Z-std, and Z-max values for the weight and bias parameters of the first and last fully connected layers are significantly elevated, which are several orders of magnitude higher than those observed in benign models. This is consistent with the high D-SNR score and indicative of model tampering introduced by the attack. For comparison, in a benign model without the attack, the z-scores remain close to zero throughout the first four warm-up steps. Even in the final step, the z-mean of \texttt{batchnorm[1]\_activation} reaches only 9.44, which remains well below the detection threshold.

%% file: concl.tex

\section{Concluding Remarks}\label{sec:concl}

This paper investigates malicious gradient leakage attacks in federated learning from a defender's perspective, focusing on safeguarding client data privacy. Through a comprehensive analysis of existing attack strategies and corresponding defenses, we uncover key insights into the nature and limitations of these threats. We demonstrate that most manually crafted model manipulations are either unrealistic in practice or readily detectable through parameter and gradient monitoring. While learning-based attacks are more sophisticated and harder to detect, they still face significant challenges when deployed in realistic FL scenarios, especially in the presence of normalization techniques commonly used in modern vision models and the federated averaging algorithm.

Our findings reveal a fundamental tension: gradient leakage attacks cannot simultaneously be highly effective at stealing private data and remain stealthy enough to avoid detection. In practice, we argue that such attacks do not constitute a serious threat to the security of FL systems. Nevertheless, to provide a comprehensive perspective, we propose a simple yet general detection mechanism capable of defending against the full spectrum of known gradient leakage attacks. Our study not only clarifies the limitations of existing threats but also offers practical, lightweight detection tools that are readily deployable in real-world federated learning settings.

%% file: manipulation.tex
\subsection{Characterization of Malicious Model Manipulation}\label{subsec:manipulation}

The success of malicious gradient leakage attacks depends on specific neural network layers where adversaries strategically introduce modifications. In this section, we analyze these targeted layers, forming the basis for detecting and mitigating such attacks.

\subsubsection{First ReLU-based fully-connected layers}

Fully-connected (FC) layers, especially those followed by ReLU activations, are common in MLP and CNN models and are prime targets for gradient leakage attacks due to their linearity. Consider a simple FC layer with ReLU activation applied to a data point $x$:
\begin{equation}
    M(x) = f(Wx + b),
\end{equation}
where $W$ represents the weights, $b$ the biases, and $f(x)$ is the ReLU activation, defined as $f(x) = \max(x, 0)$.

If this layer is located at the beginning of the network, reconstructing the input data becomes equivalent to obtaining the training data. During training, the client typically uses a mini-batch of data $\left\{x^i\right\}_{i=1}^n = X \in \mathbb{R}^{n \times m}$, where $n$ is the number of data points. The gradient of the mini-batch is the average of gradients from all data points. If the gradients for all but one data point are zero, the average gradient reduces to that of a single data point, $\mathbf{x}^*$.

In ReLU activations, the $i$-th neuron activates only if the weighted sum of features associated with negative components is less than the weighted sum of features associated with positive components:
\begin{equation}
    \sum_{n \in \mathrm{N}} w_i^{(n)} x_n + b_i^{(n)} < \sum_{p \in \mathrm{P}} w_i^{(p)} x_p + b_i^{(p)},
\end{equation}
where $N$ and $P$ represent indices for negative and positive weight components in the weight row $w_i$.

The Curious attack~\cite{curious-eurosp23} adjusts weights and biases in the first FC layer to ensure the inequality holds true for only one data point in the batch at each neuron. This isolates gradients of a single target point, making input reconstruction easier:
\begin{equation}
    x = \left(\frac{\partial \mathcal{L}}{\partial b_i}\right)^{-1} \frac{\partial \mathcal{L}}{\partial \mathbf{w}_i^T}.
\end{equation}

To amplify the signal for specific data points, the Robbing attack~\cite{robbing-iclr22} replaces the first fully-connected layer with an imprint layer, defined by a weight matrix $W \in \mathbb{R}^{k \times m}$ and bias vector $b \in \mathbb{R}^{k}$. The weights and biases are configured as $\left\langle W_i, x \right\rangle = h(x)$, where $h: \mathbb{R}^m \rightarrow \mathbb{R}$ is a linear function of the input, and $b_i = -\Phi^{-1}\left(\frac{i}{k}\right)$, with $\Phi^{-1}$ as the inverse of the standard Gaussian CDF. For image data, $h(x) = \frac{1}{m} \sum_{i=1}^m x_i$ can represent the average brightness, and the weight matrix elements are $W_{i,j} = \frac{1}{m}$.

This modification ensures gradients reflect key features of the target data, facilitating reconstruction. In the Robbing attack, the input that activates a bin is reconstructed using the gradients of weights and biases from two adjacent bins:
\begin{equation}
    x = \left(\frac{\partial \mathcal{L}}{\partial b_i} - \frac{\partial \mathcal{L}}{\partial b_{i+1}}\right)^{-1} \left(\frac{\partial \mathcal{L}}{\partial \mathbf{w}_i^T} - \frac{\partial \mathcal{L}}{\partial \mathbf{w}_{i+1}^T}\right).
\end{equation}

\textit{\textbf{Implications of modifications:}} Manipulating \textit{the first FC layer} isolates gradients for specific data points or amplifies their features, enabling input reconstruction and compromising data privacy.

\subsubsection{Convolutional layers followed by fully-connected layers}
In CNNs, FC layers typically follow several convolutional layers. While these FC layers are common targets for gradient leakage attacks, their deeper placement makes direct input reconstruction more challenging. Attackers address this by modifying the model to either bypass or simplify the convolutional layers.

One approach is to add a fully-connected layer with ReLU activation at the start of the CNN. This newly introduced FC layer corresponds directly to the input data, making it easier for attackers to exploit gradients using techniques outlined earlier. However, such architectural changes are less stealthy and easier to detect.

\begin{figure}[ht!]
    \centering
    \includegraphics[width=0.6\linewidth]{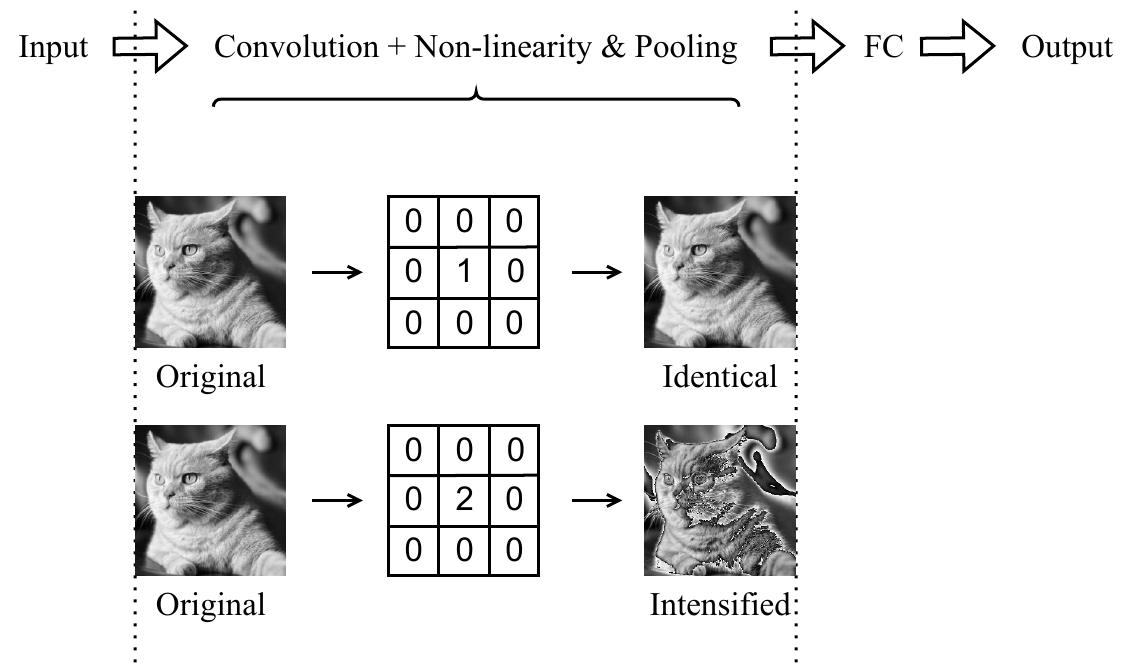}
    \caption{An example of a $3\times3$ convolutional layer with an identity kernel, which preserves the input image without any changes. Using a positive constant (\emph{e.g.,} $2$) at the center instead of $1$ scales the image features.}
    \label{fig:identity}
\end{figure}

Alternatively, attackers simplify the data flow through convolutional layers by modifying convolutional kernels to identity matrices, allowing input data to pass through unaltered~\cite{curious-eurosp23}. This ensures the input reaches subsequent FC layers without distortion or feature extraction. \cref{fig:identity} illustrates an identity convolutional kernel and its effect on transmitting 1-channel (grayscale) image data. For 3-channel (RGB) images, three identity kernels are required to propagate the original image forward through the convolutional layer. Similarly, using a positive constant (\emph{e.g.,} $2$) at the kernel's center instead of $1$ scales the input features. Reverting the scaling of the extracted data points allows for the recovery of the original data as well~\cite{curious-eurosp23}.

\textit{\textbf{Implications of modifications.}}
For CNNs, attackers aim to enable the direct flow of input data into the fully-connected layer by either \textit{inserting a new fully-connected layer at the beginning of the network to bypass the convolutional layers} or \textit{simplifying the convolutional layers to preserve the input data unaltered}.

\subsubsection{Last fully-connected classification layers}
Unlike attacks targeting layers closer to the input, the Fishing attack~\cite{fishing-icml22} manipulates the last classification layer to isolate gradient updates associated with a single class or feature. This makes it easier to reverse-engineer and reconstruct the corresponding data.

To achieve this, the Fishing attack modifies the weights and biases of the last fully-connected layer. For a target class $c$, if there is only one image in the class, the weight matrix $W \in \mathbb{R}^{p \times q}$ and bias vector $b \in \mathbb{R}^{q}$ can be configured to focus exclusively on that image, for example:
\begin{equation}
    \begin{aligned}
        W_{i, j} & = \begin{cases} W_{i, j}, & \text{if } i = c \\ 
        0, & \text{otherwise} \end{cases} \\ 
        b_i & = \begin{cases} b_i, & \text{if } i = c \\ 
        \alpha, & \text{otherwise} \end{cases}
    \end{aligned},
    \label{eq:fishing}
\end{equation}
where $\alpha$ and $\theta$ are hyperparameters for controlling the relative magnitude of the gradient signal associated with the target data.



\textit{\textbf{Implications of modifications.}}
For CNNs, in addition to targeting earlier layers that process raw input data, attackers can also manipulate the weights and biases of \textit{the last classification layer} to reduce the mixed gradient updates to those associated with a single class or feature.

\subsubsection{Transformers}
In Transformers, attackers face challenges reconstructing input data due to the complex mixing of token information across layers, making it difficult to isolate and recover input tokens from gradient updates. The Decepticons attack~\cite{decepticons-iclr23} addresses this by disabling all attention layers and restricting the outputs of feed-forward blocks to the last entry. This ensures token information remains unmixed, keeping inputs to each layer unaltered.

Each attention layer operates as described in~\cref{eq:atten}, where $Q$, $K$, and $V$ represent query, key, and value matrices derived from input embeddings~\cite{transformers-emnlp20}. The attack modifies the attention mechanism as shown in~\cref{eq:mali_atten}: key weights $W_K$ are set to identity matrices, key biases $b_K$ are set to zero, query weights $W_Q$ are zero, and query biases $b_Q$ are scaled versions of the first positional encoding $\vec{p}_0$. Value weights $W_V$ are partial identity matrices controlled by a hyperparameter $d'$, and value biases $b_V$ are zero.
\begin{equation}
    \begin{aligned}
        \text{Attention}(Q, K, V) & = \text{Softmax}\left(\frac{Q K^\top}{\sqrt{d_k}}\right) V \\
    \end{aligned}
    \label{eq:atten}
\end{equation}
\begin{equation}
    \begin{aligned}
        W_K = \mathbf{1}_{d_{\text{model}}}, & \;\;\; b_K = \mathbf{0},             \\
        W_Q = \mathbf{0},                    & \;\;\; b_Q = \gamma \cdot \vec{p}_0, \\
        W_V = \mathbf{1}_{d'},               & \;\;\; b_V = \mathbf{0}.
    \end{aligned}
    \label{eq:mali_atten}
\end{equation}

Combined with this modification, the Decepticons attack~\cite{decepticons-iclr23} further applies techniques from the Robbing attack~\cite{robbing-iclr22} to modify the weight and bias of the first linear layer with activation in each Transformer block.

\textit{\textbf{Implications of modifications.}}
For Transformers, attackers aim to bypass \textit{attention and feed-forward layers}, ensuring token information remains unmixed and is directly forwarded to the fully-connected layer for easier extraction.